\documentclass[letterpaper, 10 pt, conference]{ieeeconf}

\IEEEoverridecommandlockouts

\usepackage{cite}
\usepackage{amsmath,amssymb,amsfonts}
\usepackage{graphicx}
\usepackage{booktabs}
\usepackage{tabularx}
\usepackage{hyperref}

\newcommand{\pairW}{0.45\textwidth}
\newcommand{\pairH}{0.22\textheight}
\def\BibTeX{{\rm B\kern-.05em{\sc i\kern-.025em b}\kern-.08em
    T\kern-.1667em\lower.7ex\hbox{E}\kern-.125emX}}

\title{\LARGE \bf
AutoClimDS: Climate Data Science Agentic AI \\-- A Knowledge Graph is All You Need}
\author{
Ahmed Jaber$^{1,*}$\thanks{Corresponding author. Email: \href{mailto:amj2234@columbia.edu}{amj2234@columbia.edu}},
Wangshu Zhu$^{1}$,
Ayon Roy$^{1}$,
Candace Agonafir$^{1}$,
Linnia Hawkins$^{1}$,
\\[0.3em]
Karthick Jayavelu$^{2}$,
Justin Downes$^{2}$,
Sameer Mohamed$^{2}$,
Tian Zheng$^{1,3,*}$\thanks{Corresponding author. Email: \href{mailto:tian.zheng@columbia.edu}{tian.zheng@columbia.edu}}\thanks{Code and supplementary materials: \href{https://github.com/Ajaberr/AutoClimDS}{https://github.com/Ajaberr/AutoClimDS}}\thanks{This research is supported by NSF through the Learning the Earth with Artificial intelligence and Physics (LEAP) Science and Technology Center (STC) (Award \#2019625).}
\\[0.6em]
$^{1}$NSF STC Learning the Earth with AI and Physics (LEAP), Columbia University, New York, NY, USA;\\
$^{2}$AWS Generative AI Innovation Center, Seattle, WA, USA;\\
$^{3}$Department of Statistics, Columbia University, New York, NY, USA;\\
$^{*}$ Corresponding authors.}

\begin{document}

\maketitle

\thispagestyle{empty}
\pagestyle{empty}

\begin{abstract}
Climate data science remains constrained by fragmented data sources, heterogeneous formats, and steep technical expertise requirements. These barriers slow discovery, limit participation, and undermine reproducibility. We present \textit{AutoClimDS}, a Minimum Viable Product (MVP) Agentic AI system that addresses these challenges by integrating a curated climate knowledge graph (KG) with a set of Agentic AI workflows designed for cloud-native scientific analysis. The KG unifies datasets, metadata, tools, and workflows into a machine-interpretable structure, while AI agents, powered by generative models, enable natural-language query interpretation, automated data discovery, programmatic data acquisition, and end-to-end climate analysis. A key result is that AutoClimDS can reproduce published scientific figures and analyses from natural-language instructions alone, completing the entire workflow from dataset selection to preprocessing to modeling. When given the same tasks, state-of-the-art general-purpose LLMs (e.g., ChatGPT GPT-5.1) cannot independently identify authoritative datasets or construct valid retrieval workflows using standard web access. This highlights the necessity of structured scientific memory for agentic scientific reasoning. By encoding procedural workflow knowledge into a KG and integrating it with existing technologies (cloud APIs, LLMs, sandboxed execution), AutoClimDS demonstrates that the KG serves as the essential enabling component, the irreplaceable structural foundation, for autonomous climate data science. This approach provides a pathway toward democratizing climate research through human–AI collaboration.
\end{abstract}


\begin{center}
\footnotesize
© 2026 IEEE. Personal use of this material is permitted. Permission from IEEE must be obtained for all other uses, in any current or future media, including reprinting/republishing this material for advertising or promotional purposes, creating new collective works, for resale or redistribution to servers or lists, or reuse of any copyrighted component of this work in other works.
\end{center}

\section{INTRODUCTION}
Climate research investigates Earth's climate systems, variability, and change effects on natural and human environments \cite{eyring2024pushing}, drawing from observational datasets, simulation outputs, and analytical tools across domains \cite{gettelman2022future}. Despite data proliferation, research remains fragmented: datasets in heterogeneous formats with inconsistent metadata lack standardized access \cite{ceccato2012data}. Existing retrieval systems rely on keyword search requiring users to know dataset names \cite{shum2017harvesting}, while general-purpose LLMs lack the structured scientific memory needed for autonomous data acquisition and workflow construction. Importantly, AutoClimDS does not position the knowledge graph as a replacement for large language models or computational tools, but as the structural substrate that enables their reliable coordination. In this context, the phrase “a knowledge graph is all you need” should be read as a statement about necessity rather than sufficiency: without a structured scientific memory encoding datasets, access paths, and procedural knowledge, even frontier LLMs fail to perform autonomous scientific workflows.

\textbf{Contributions:} We introduce a semantic infrastructure encoding climate data entities into a unified, queryable knowledge graph (KG) serving as structured scientific memory for agentic workflows. First, we present an ontology-driven methodology integrating NASA CMR, NOAA OneStop \cite{noaa_onestop}, ERA5 \cite{hersbach2020era5}, and CMIP6 \cite{eyring2016cmip6} records into a semantically consistent graph using OpenCypher \cite{green2018opencypher}. Unlike prior KGs providing conceptual vocabularies or keyword-driven retrieval, this approach encodes procedural workflow knowledge: executable access links with authentication protocols, variable-level semantic mappings via fine-tuned ClimateBERT \cite{webersinke2021climatebert} (99.17\% accuracy), geospatial relationships, and preprocessing operation metadata. Second, we demonstrate this KG as a reasoning substrate for \textit{AutoClimDS}, reproducing scientific workflows end-to-end where existing approaches fail. The KG provides essential structural constraints, typed relationships, spatial containment, irreplaceable by pure vector search or LLM reasoning, evidenced by successful replication of published climate studies (Section III).

\subsection{Agentic AI and Human–AI Collaboration}

Agentic AI systems \cite{acharya_agentic_2025} capable of autonomous reasoning, planning, and tool use offer promising solutions. Evidence shows generative AI assistants boost productivity 14\% on average, with the greatest gains for less-experienced users \cite{workersAI}, narrowing expertise gaps in complex tasks. In climate data science, where bottlenecks stem from technical barriers, knowledge-graph-enabled Agentic AI could elevate entry-level researchers while streamlining expert workflows, making discovery more inclusive and reproducible.

\subsection{Related Work}

Several initiatives have sought to improve climate data access and interoperability. Foundational projects like NASA Earthdata \cite{nasaearthdata}, CMIP \cite{meehl2000coupled}, and ESGF \cite{williams2011earth} provide observational and model datasets. Computational tools such as Pangeo \cite{odaka2019pangeo} and ESMValTool \cite{righi2020earth} facilitate cloud-based analysis, while ontologies like SWEET \cite{raskin2005knowledge} and GeoLink \cite{zhou2020geolink} offer structured vocabularies for Earth science concepts. However, these were designed for human researchers: SWEET provides conceptual hierarchies (e.g., "Precipitation" $\subseteq$ "AtmosphericPhenomenon") but lacks executable data access paths; Pangeo requires manual dataset specification; ESMValTool demands expert-configured recipes. None capture the procedural reasoning needed for autonomous agentic workflows, how to authenticate with NASA Earthdata, which variable to extract from multi-dimensional NetCDF, or which pre-processing operations to apply for specific analyses.

Recent efforts, such as LinkClimate \cite{wu2022linkclimate} demonstrated knowledge graph infrastructures for climate datasets. However, its design reflects a static view where discovery is keyword-driven, treating datasets as retrieval endpoints without representing procedural knowledge or scientific reasoning needed for workflow construction. When tasked with reproducing the NPCC4 sea level analyses we used to illustrate AutoClimDS (Section III), LinkClimate's keyword search returned dataset titles but relied on exact string matching over relatively shallow metadata fields. Similarly, in our experiments, GPT-5.1 with web search but without KG guidance failed to autonomously locate authoritative datasets, instead hallucinating dataset names or selecting inappropriate sources with mismatched temporal/spatial coverage. These experiment results (can be found in the AutoClimDS \href{https://github.com/Ajaberr/AutoClimDS}{GitHub repo}) reveal a critical gap: the lack of infrastructure serving as a reasoning backbone for Agentic AI. The current paper addresses this gap by developing a knowledge graph encoding not just data locations but procedural reasoning paths, demonstrating that structured scientific memory enables reliable agentic workflows where pure retrieval or LLM reasoning alone fails.

\section{METHOD}

\subsection{Knowledge Graph Ontology and Construction}

The knowledge graph integrates metadata from NASA CMR, NOAA OneStop \cite{noaa_onestop}, ERA5 \cite{hersbach2020era5}, and CMIP6 \cite{eyring2016cmip6} via respective APIs. The graph contains $|\mathcal{V}| \approx 1{,}480{,}000$ nodes across 45 types and $|\mathcal{E}| \approx 5{,}800{,}000$ edges across 39 relationship types, encompassing $\sim$208,000 climate datasets (106,000 observational, 102,000 simulation outputs). The schema partitions into two branches: observational data (NASA CMR/NOAA OneStop capturing measurements, satellite retrievals, reanalysis) and simulation data (CMIP6/ERA5 encoding experiments, ensembles, model outputs). This hybrid design encodes both data content and procedural context.

\textbf{Data ingestion pipeline:} NASA CMR collections and granules were retrieved via Search API (\texttt{cmr.earthdata.nasa.gov}) with UMM-JSON parsing for temporal/geospatial metadata. NOAA OneStop (\texttt{data.noaa.gov/onestop}) integrates NCEI, ERDDAP, CO-OPS repositories with a 5 req/s rate limiting. ERA5 datasets were discovered via Copernicus CDS web crawling (\texttt{cds.climate.copernicus.eu}) with HTML parsing and JSON normalization. CMIP6 metadata resolves via ESGF distributed index API with 10-field DRS tuple filtering (mip\_era, activity\_id, institution\_id, source\_id, experiment\_id, variant\_label, table\_id, variable\_id, grid\_label, version) and multi-node failover.

\textbf{Geospatial processing:} Polygon coordinates from dataset metadata undergo Shapely geometric processing for bounding box extraction. Mapbox Geocoding API classifies spatial scope hierarchically: ocean, global, continental, country, multinational, or regional across 258 location boundaries. This enables spatial relationship traversal via \texttt{hasLocation} edges during multi-criteria search.

\textbf{Link scoring and validation:} URLs undergo automated downloadability assessment. Domain analysis identifies API types (OpenDAP, ERDDAP, WMS, WFS, REST) and authentication requirements (public vs. restricted access). Downloadability weights reflect empirical retrieval success on an integer scale: direct download links ($w=5$), data access portals ($w=4$), offline access endpoints ($w=3$), visualization services ($w=2$), informational links ($w=1$), with verified endpoints receiving an additional reliability bonus ($+1$). Endpoint verification tests validate data accessibility before graph ingestion.


{\raggedright
\textbf{Semantic variable mapping:} We fine-tuned ClimateBERT \cite{webersinke2021climatebert} \footnote{\texttt{climatebert/distilroberta-base-climate-f}} for mapping natural language inquiries to climate variables through multi-class classification over 2,308 CESM variables. Architecture employs attention-masked mean pooling with dropout ($p=0.3$), trained via cross-entropy loss with Adam optimizer ($\alpha = 1 \times 10^{-5}$, batch 16, 50 epochs) on curated CESM training set. Similarity-based clustering addresses variable redundancy via a similarity function $S(\text{desc}(v_i), \text{desc}(v_j)) \geq 0.7$, where $S$ denotes the cosine similarity between textual description embeddings. This process achieves an exact match accuracy of $\mathcal{A}_{\text{exact}} = 93.45\%$ and a group-level accuracy of $\mathcal{A}_{\text{group}} = 99.87\%$. The trained model maps observational dataset variables to standardized CESM nomenclature, enabling cross-dataset variable discovery.
\par}

\textbf{OpenCypher graph construction:} JSON metadata converts to Neptune-compatible OpenCypher CSV format via schema-driven transformation. Node types partition into core entities and climate-simulation-specific types. Relationships encode typed edges with source-target node constraints. Embeddings for vector-enabled nodes are generated via sentence-transformers. The complete schema with node/edge definitions is available in the supplementary materials \cite{autoclimds}.

\subsection{AutoClimDS Agentic AI Architecture}

The system implements three core objectives: \textit{Data Discovery}, \textit{Data Acquisition}, and \textit{Climate Modeling/Analytics}. Implementation uses LangChain \cite{LangGraph} with ReAct reasoning \cite{yao2023react} and Bedrock Claude Sonnet 4 \cite{aws_bedrock_claude_sonnet_4}.

\subsubsection{Data Discovery Agent}

The agent implements semantic dataset discovery by encoding research queries into 384-dimensional vectors through sentence-transformers. Vector search operates through Neptune Analytics' \texttt{topKByEmbedding()} procedure, implementing hierarchical navigable small world graphs for efficient approximate nearest neighbor retrieval \cite{AWSNeptuneAnalytics}.

The agent implements intelligent search routing through embedding availability detection, categorizing node types into vector-enabled and text-only categories. Vector-enabled types include \texttt{DataCategory}, \texttt{Variable}, \texttt{CESMVariable}, \texttt{ScienceKeyword}, \texttt{Location}, \texttt{TemporalResolution}, and \texttt{SpatialResolution}, while remaining nodes utilize text-based Neptune query matching.

Multi-criteria search extends single-criterion functionality by combining vector results with relationship-based filtering, addressing real-world complexity where researchers must consider temporal coverage, spatial resolution, variable availability, and institutional provenance simultaneously. The algorithm constructs complex OpenCypher queries incorporating vector results as node constraints while applying additional filtering through temporal overlap detection $[t_{\texttt{start}}, t_{\texttt{end}}]$, spatial relationship traversal via \texttt{hasLocation} relationships, and other edges. This hybrid approach enables queries like "precipitation datasets over the Pacific Northwest from 1980-2020 with daily resolution" to leverage vector search for "precipitation" while applying structured filters for location, temporal range, and resolution metadata.

Retrieved datasets undergo link reranking for automated prioritization: for dataset $D$ with links $L_D = \{l_1, \ldots, l_n\}$, each link $l_i$ possesses preprocessed weight $w_i \in \{0,1,2,3,4,5\}$ assigned during graph construction. The reranking function orders by descending weight via permutation $\sigma$: $l_{\sigma(1)}, \ldots, l_{\sigma(n)}$ where $w_{\sigma(1)} \geq \cdots \geq w_{\sigma(n)}$. This prioritization ensures direct download links ($w=5$) are attempted before data access portals ($w=4$), offline access endpoints ($w=3$), visualization services ($w=2$), or informational links ($w=1$), with verified endpoints receiving an additional reliability bonus ($+1$). The reranking operation executes as post-processing following dataset retrieval, transforming unordered link collections into priority-ordered access sequences, maximizing the probability of successful automated data acquisition. Results persist to local SQLite database, enabling session continuity, recall of previously retrieved datasets, and support for iterative research workflows.

\subsubsection{Data Acquisition Agent}


Once the knowledge graph returns relevant datasets with reranked access links, the Agentic AI system transitions from discovery to acquisition. Each dataset entry in the research database is inspected, and the agent retrieves corresponding links from \texttt{hasLink} relationships of dataset nodes, pre-ordered by downloadability weight. Based on research query $q$, the agent determines next action $a \in \mathcal{A}$ where the action space is defined as $\mathcal{A} = \{\text{retrieve}, \text{preprocess}, \text{analyze}\}$.

If $a = \texttt{retrieve}$, the agent first attempts data acquisition through the highest-weighted links. When links provide explicit API endpoints or direct download URLs, the agent invokes the appropriate retrieval protocol with authentication handled via preconfigured tokens (e.g., NASA Earthdata credentials, NOAA CDO API keys). However, when links lack clear programmatic access methods or when initial retrieval attempts fail, the agent engages in dynamic access discovery. This process leverages web search and website fetching tools to query documentation, data portals, and technical specifications, enabling the agent to discover access protocols autonomously rather than relying on hardcoded patterns. The agent can read API documentation, identify authentication requirements, locate endpoint specifications, and generate custom retrieval code adapted to the specific data source. For NOAA datasets requiring location-based queries, the agent utilizes a location code resolution API with token validation to translate geographic descriptors into standardized location identifiers before constructing data requests.

Raw data, denoted $D = \{d_1, d_2, \ldots, d_n\}$, may arrive in heterogeneous formats such as CSV, NetCDF, HDF, or JSON. An automated transformation function $T: D \mapsto \hat{D}$ standardizes the collection into tabular or array-based structures, enabling interoperability. Quality validation steps are expressed as a constraint-checking function $V(\hat{D}) \in \{0,1\}$, which enforces link validity, accessibility, and structural consistency. Only datasets satisfying $V(\hat{D})=1$ are retained for downstream workflows. These steps execute through \texttt{CodeExecutionTool} operating within sandboxed Python execution environments with controlled namespace isolation, network access authenticated against whitelisted repositories (NASA Earthdata, NOAA, AWS S3), and output sanitization. The adaptive pipeline enables protocol discovery for previously unseen data sources.

By integrating discovery, retrieval, validation, and preprocessing into a single agent-driven workflow, the system achieves reproducible and cloud-resilient data acquisition. Graph linkages to cloud repositories ensure persistence, while transformation $T$ and validation $V$ guarantee dataset usability. The agent's dynamic protocol discovery ensures robustness across a heterogeneous climate data infrastructure where access methods vary significantly.

\subsubsection{Climate Simulation Agents}

Two specialized agents handle ERA5/CMIP6 datasets. \textit{Discovery agent} executes OpenCypher queries against Neptune for \texttt{SimDataset} nodes, leveraging relationships. For CMIP6, DRS filtering uses institution/scenario/ensemble relationships. Link validation confirms CDS API (ERA5) or ESGF HTTP (CMIP6) accessibility. \textit{Acquisition agent} invokes \texttt{cdsapi} for ERA5 asynchronous retrieval with authentication, queuing, polling. For location queries (e.g., "NYC temperature"), geocoding resolves descriptors to bounding boxes for spatial subsetting,a query for "New York City SSP2-4.5 temperature projections" automatically translates to coordinates [40.7°N, -74.0°W] with an appropriate buffer for spatial subsetting of global model grids. CMIP6 uses authenticated HTTP from ESGF nodes prioritized by link weights. Downloaded NetCDF/GRIB files undergo \texttt{xarray}-based loading with specialized tools (\texttt{ProcessERA5DataTool}, \texttt{ProcessCMIP6DataTool}, \texttt{TimeSeriesAnalysisTool}, \texttt{SpatialAnalysisTool}, \texttt{CompareERA5CMIP6Tool}) handling calendar conventions, coordinate transforms, and unit conversions.

\subsubsection{State Management and Error Recovery}

The system implements persistent state management via SQLite for dataset discovery results and LangChain's \texttt{ConversationBufferWindowMemory} ($k=15$ exchanges for Orchestrator, $k=5$ for specialized agents), where $k$ denotes the maximum number of previous conversation exchanges maintained in the active buffer. When data acquisition fails, cascading fallback executes: (1) iterate through reranked links $l_{\sigma(1)}, \ldots, l_{\sigma(n)}$ from Section 2.2.1, (2) invoke dynamic discovery via web search/documentation fetching when all pre-indexed links fail, (3) traverse \texttt{hasCESMVariable} relationships to identify alternative datasets containing the same standardized climate variables if the target dataset remains inaccessible. \texttt{CodeExecutionTool} operates within sandboxed Python execution with controlled namespace isolation, enabling multi-step preprocessing through LangChain's conversation memory.

\subsubsection{Agentic Loop Mechanics and Guardrails}

The system implements ReAct (Reasoning + Acting) \cite{yao2023react} through iterative cycles. For query $q$, the loop executes: $\text{Thought}_t = \texttt{LLM}(q, \mathcal{H}_{t-1}, \mathcal{O}_{t-1})$, $\text{Action}_t = \texttt{parse}(\text{Thought}_t) \in \mathcal{A}$, $\mathcal{O}_t = \texttt{Tool}(\text{Action}_t)$, $\mathcal{H}_t = \mathcal{H}_{t-1} \cup \{(\text{Thought}_t, \text{Action}_t, \mathcal{O}_t)\}$, where $\mathcal{H}_t$ is conversation history, $\mathcal{O}_t$ is tool observation, and $\mathcal{A}$ is the action space.

To prevent infinite loops and token exhaustion, agents enforce: (1) \texttt{max\_iterations}=15 for discovery/acquisition agents, 100 for Orchestrator handling complex workflows, (2) \texttt{handle\_parsing\_errors}=True for malformed LLM outputs, (3) \texttt{max\_execution\_time}=300s timeout, (4) semantic cycle detection terminating when consecutive thought embeddings exceed cosine similarity 0.95 for three steps, (5) circuit breaker for identical tool calls within 5-step window. Multi-agent workflows use LangChain's \texttt{StateGraph} defining directed acyclic graphs with explicit routing logic: Orchestrator → Discovery Agent → Acquisition Agent, with conditional fallback edges routing back to Discovery upon acquisition failure rather than allowing indefinite Acquisition loops.

\subsubsection{Multi-Agent System}

\begin{figure}[!t]
\centering
\includegraphics[width=0.9\columnwidth]{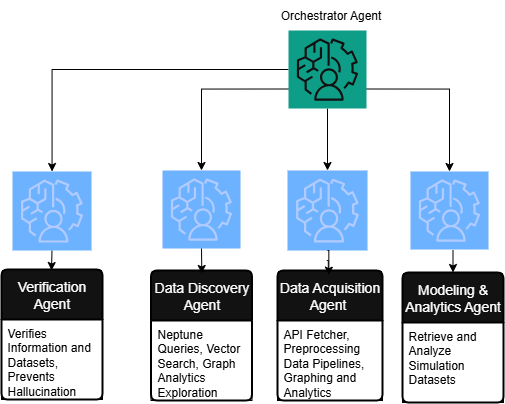}
\caption{Multi-agent system architecture.}
\label{fig:multiagent-arch}
\end{figure}

The architecture (Fig.~\ref{fig:multiagent-arch}) features a central Orchestrator Agent interpreting objectives, maintaining session state via persistent storage (Section 2.2.5), and delegating to specialized agents through \texttt{StateGraph}-based workflow execution (Section 2.2.6). Data Discovery Agent queries the KG, Data Acquisition Agent retrieves from cloud sources with automated link fallback and error recovery, and Climate Modeling Agent integrates model ensembles. When validation fails ($V(\hat{D})=0$), the system logs failure reason and routes back to Discovery for alternative datasets, ensuring transparency and preventing silent failures. When the Acquisition Agent exhausts all links, the workflow routes to Discovery for alternatives rather than halting, ensuring robustness through automated fault tolerance.

\subsection{Cloud Deployment}

\begin{figure*}[h]
\centering
\includegraphics[width=0.9\textwidth]{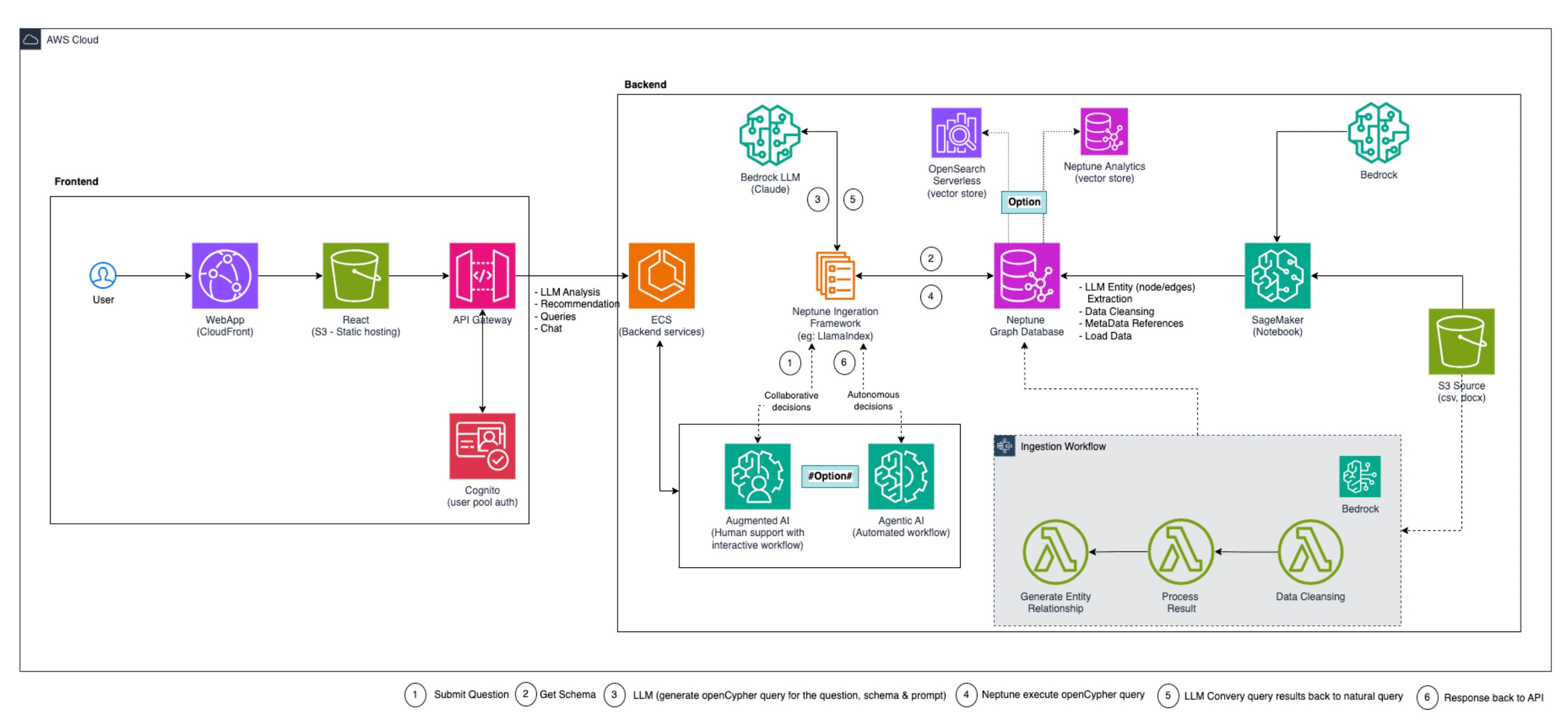}
\caption{End-to-end AWS architecture with frontend (CloudFront, React, API Gateway, Cognito) and backend (Bedrock, Neptune, SageMaker) integration.}
\label{fig:aws}
\end{figure*}

The reference implementation deploys on AWS Neptune for graph storage and Bedrock for LLM inference (Fig.~\ref{fig:aws}). However, the KG schema and agent workflows are cloud-agnostic: the 45-node-type, 39-relationship-type schema exports to OpenCypher CSV format compatible with Neo4j, ArangoDB, or any OpenCypher graph database; agents can use open-source models (Llama, Mistral) via HuggingFace; vector search can use FAISS or Qdrant. The scientific contribution lies in the KG's encoding of procedural climate data science workflows, not the cloud infrastructure. The details of the implementation and the computational costs are documented in the supplementary materials \cite{autoclimds}.

\subsection{Limitations}
AutoClimDS is a system whose performance depends on the coverage and quality of its underlying knowledge graph. Dataset discovery is currently limited to sources indexed from NASA CMR, NOAA OneStop, ERA5, and CMIP6, and datasets outside these repositories require manual graph extension. While link validation and reranking improve robustness, some data sources impose authentication, rate limits, or intermittent availability that can disrupt fully autonomous execution. Additional implementation constraints and known failure cases are documented in the project repository and supplementary materials \cite{autoclimds}.

\section{CASE STUDIES}

\subsection{Observational Data: Sea Level Trends}

We replicate figures from \textit{NYC Climate Risk Information 2022 (NPCC4)} \cite{braneonNPCC4NewYork2024} to validate AutoClimDS' capability for reproducing climate risk indicators using natural language queries. Data discovery and acquisition proceeded via natural language instructions only: no datasets, numerical values, or coefficients are provided. The instructions followed a three-part structure: (1) \textit{Objective}, (2) \textit{Context/Constraints}, (3) \textit{Desired Output}. Agents autonomously located datasets, preprocessed, calculated measures, generated figures.

\begin{figure*}[!t]
\centering
\begin{tabular}{cc}
\includegraphics[width=\pairW,height=\pairH]{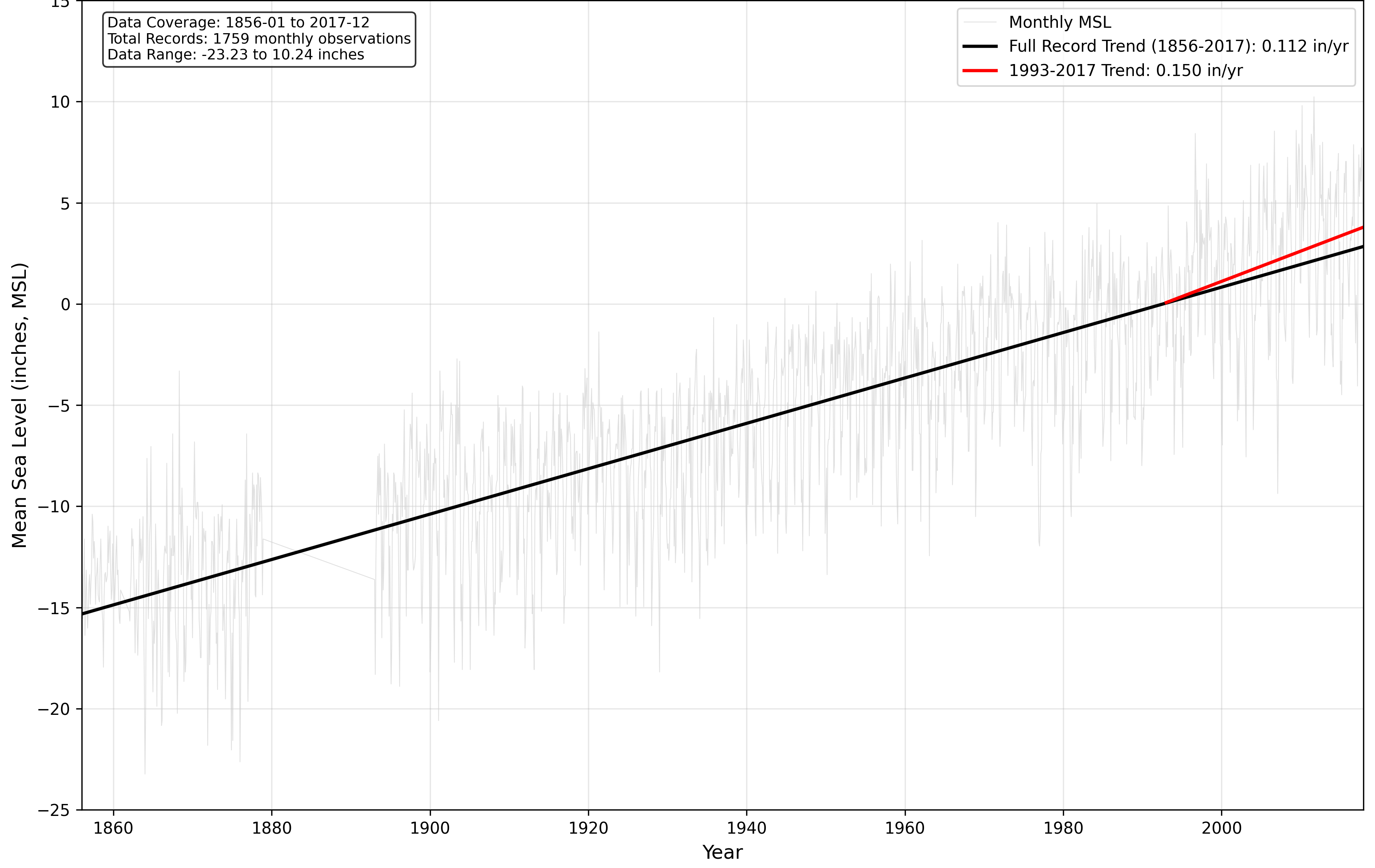} &
\includegraphics[width=\pairW,height=\pairH]{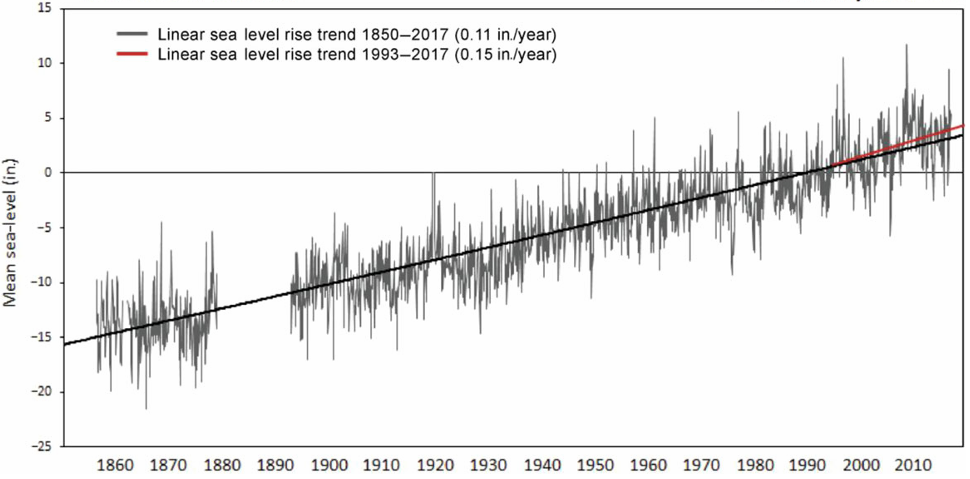} 
\end{tabular}
\caption{AutoClimDS replicated NPCC4 sea level trends.}
\label{fig:reproduced-npcc4}
\end{figure*}

\begin{figure*}[!t]
\centering
\begin{tabular}{cc}
\includegraphics[width=\pairW,height=\pairH]{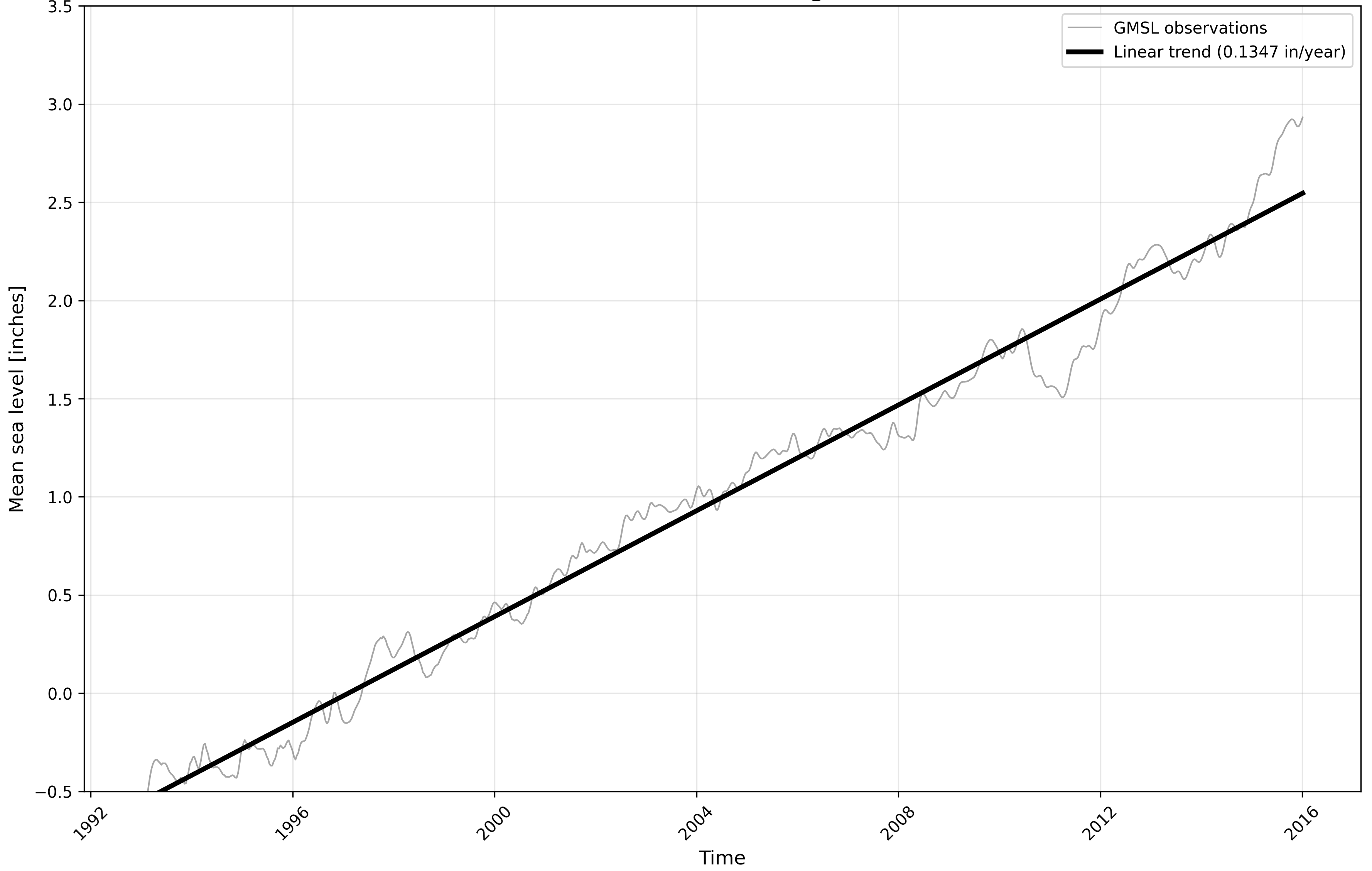} &
\includegraphics[width=\pairW,height=\pairH]{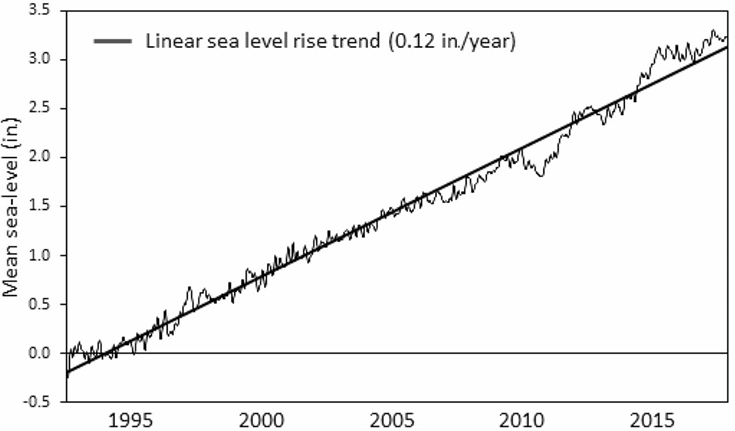} 
\end{tabular}
\caption{Original figures from \cite{braneonNPCC4NewYork2024} (CC BY-NC license).}
\label{fig:original-npcc4}
\end{figure*}

\begin{figure*}[!t]
\centering
\begin{tabular}{cc}
\includegraphics[width=\pairW,height=\pairH]{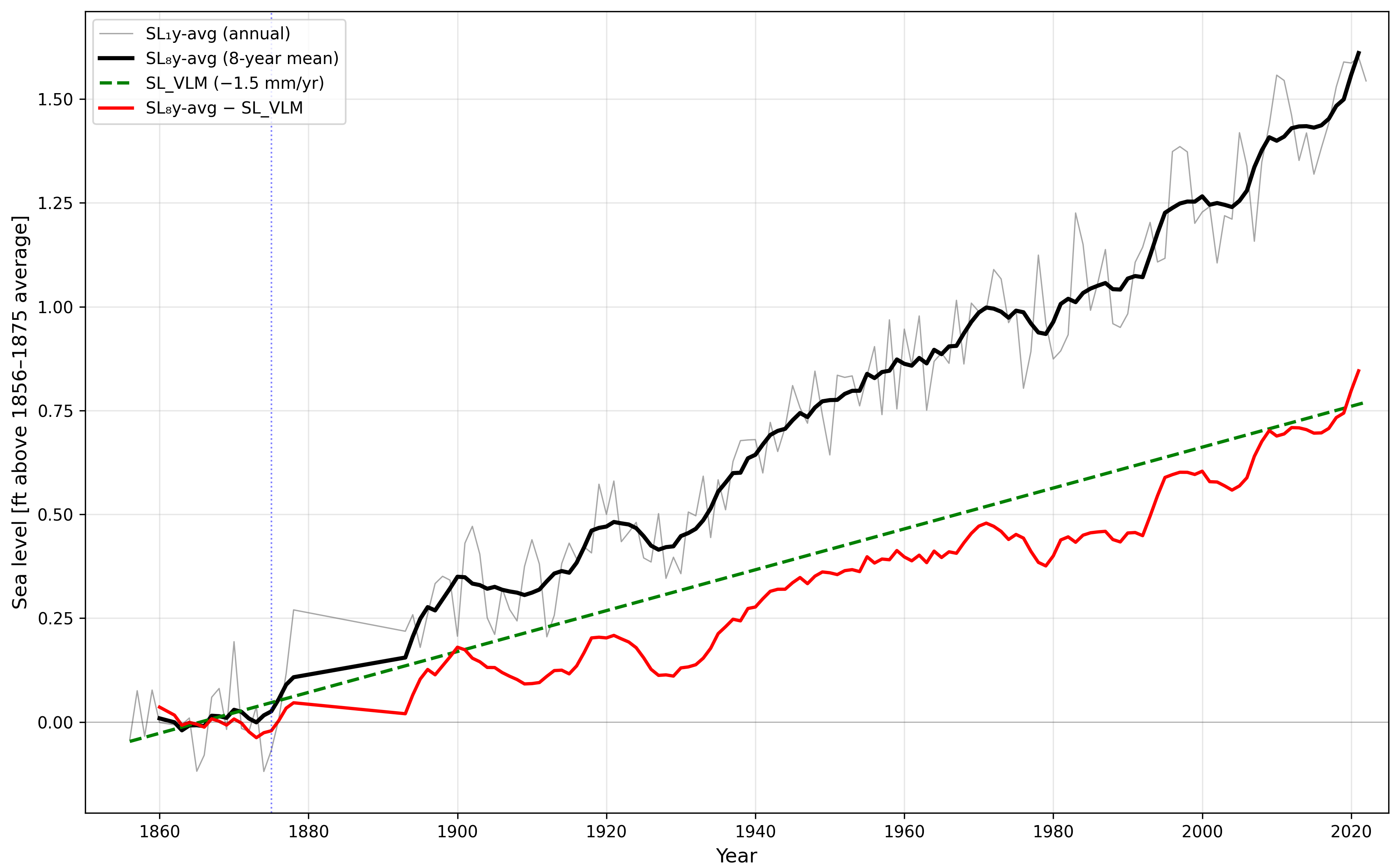} &
\includegraphics[width=\pairW,height=\pairH]{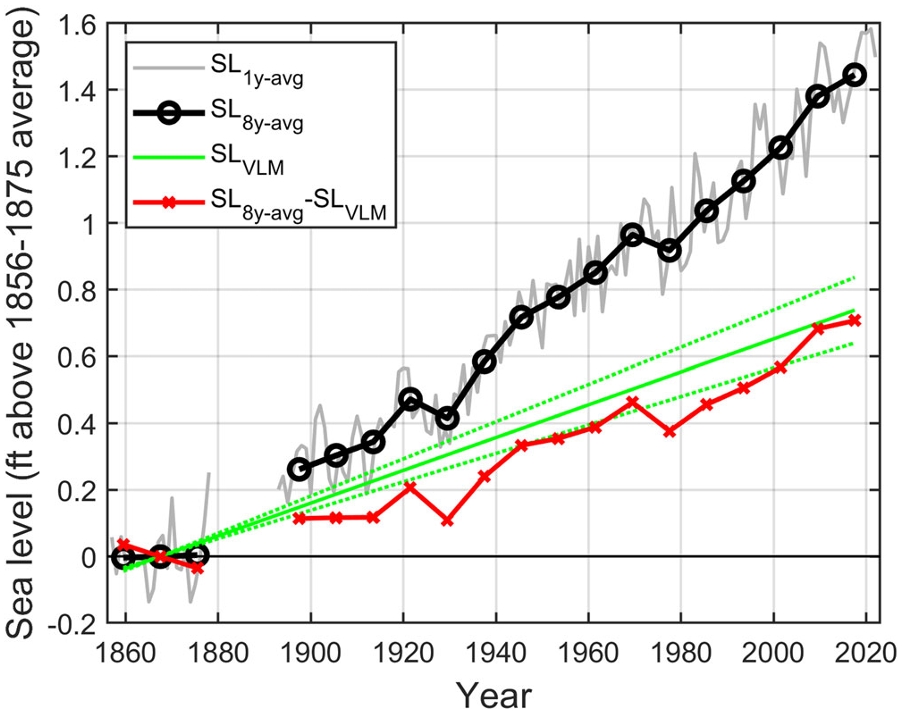}
\end{tabular}
\caption{Sea level trends with VLM-driven SLR: AutoClimDS (left) vs.\ Original \cite{braneonNPCC4NewYork2024} (right).}
\label{fig:main}
\end{figure*}

\textbf{Baseline Comparisons:} As established in Section 1.2, existing approaches (LinkClimate, GPT-5.1, NASA CMR) failed to complete these workflows autonomously. In contrast, AutoClimDS successfully replicated all NPCC4 figures (Figs.~\ref{fig:reproduced-npcc4}--\ref{fig:main}). Statistical validation confirms high fidelity across all indicators. For Battery Park sea level, AutoClimDS calculates a long-term trend of 0.112 in/yr, exceeding the precision of the reported 0.11 in/yr, and exactly reproduces the recent acceleration of 0.150 in/yr (1993–2017). Similarly, the system accurately recovers the Vertical Land Motion (VLM) contribution of $-1.5$ mm/yr and the Global Mean Sea Level (GMSL) trend of 0.12 in/yr. Additionally, the Jensen-Shannon Divergence (JSD) is 0, confirming that the graph pairs are exactly the same. These metrics demonstrate AutoClimDS reproduces published analyses with statistical equivalence. All logs, prompts, data, and figures available in supplementary materials \cite{autoclimds}.

\subsection{Climate Simulation: Temperature Projections}

\begin{figure*}[!t]
\centering
\begin{tabular}{cc}
\includegraphics[width=\pairW]{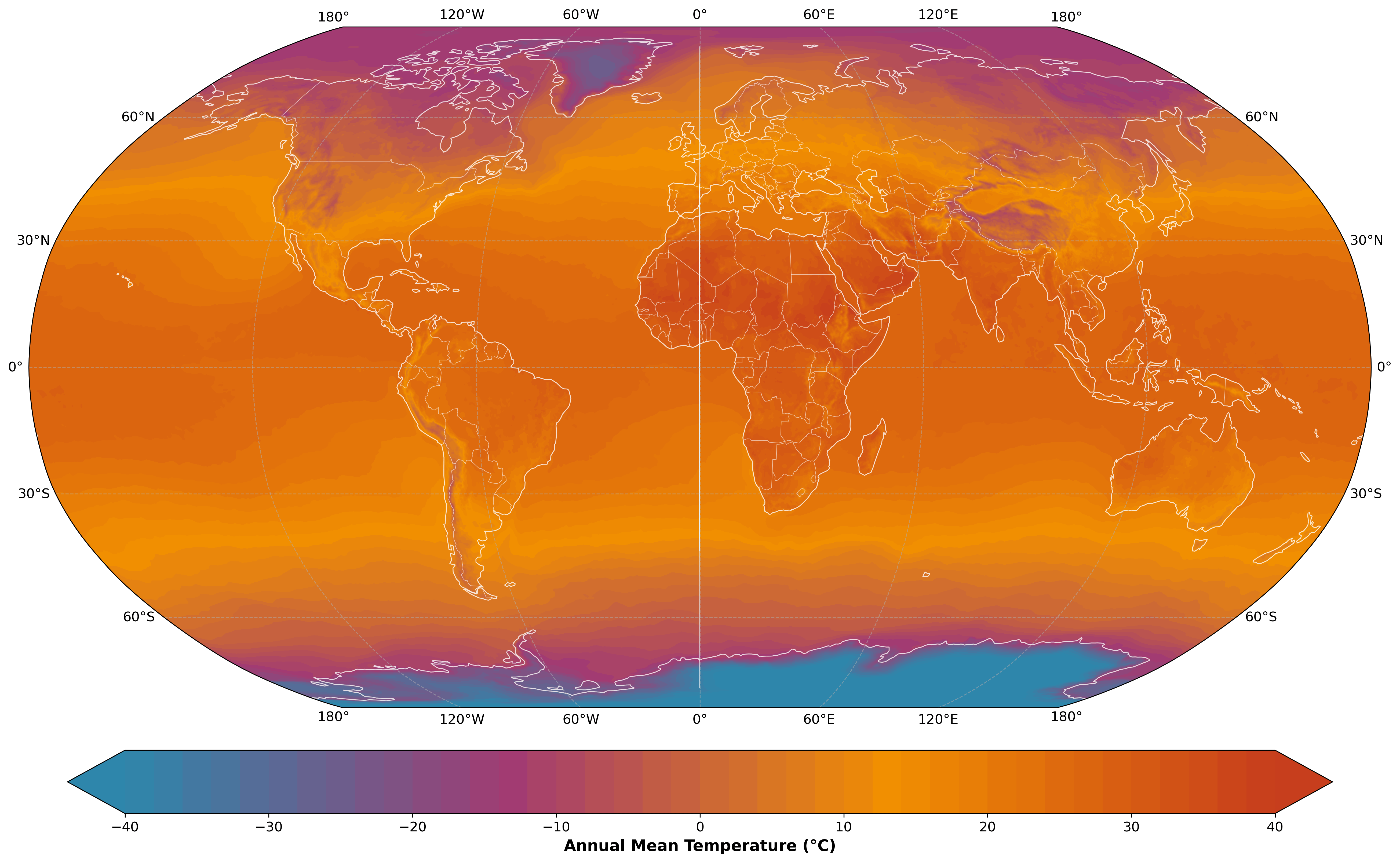} &
\includegraphics[width=\pairW]{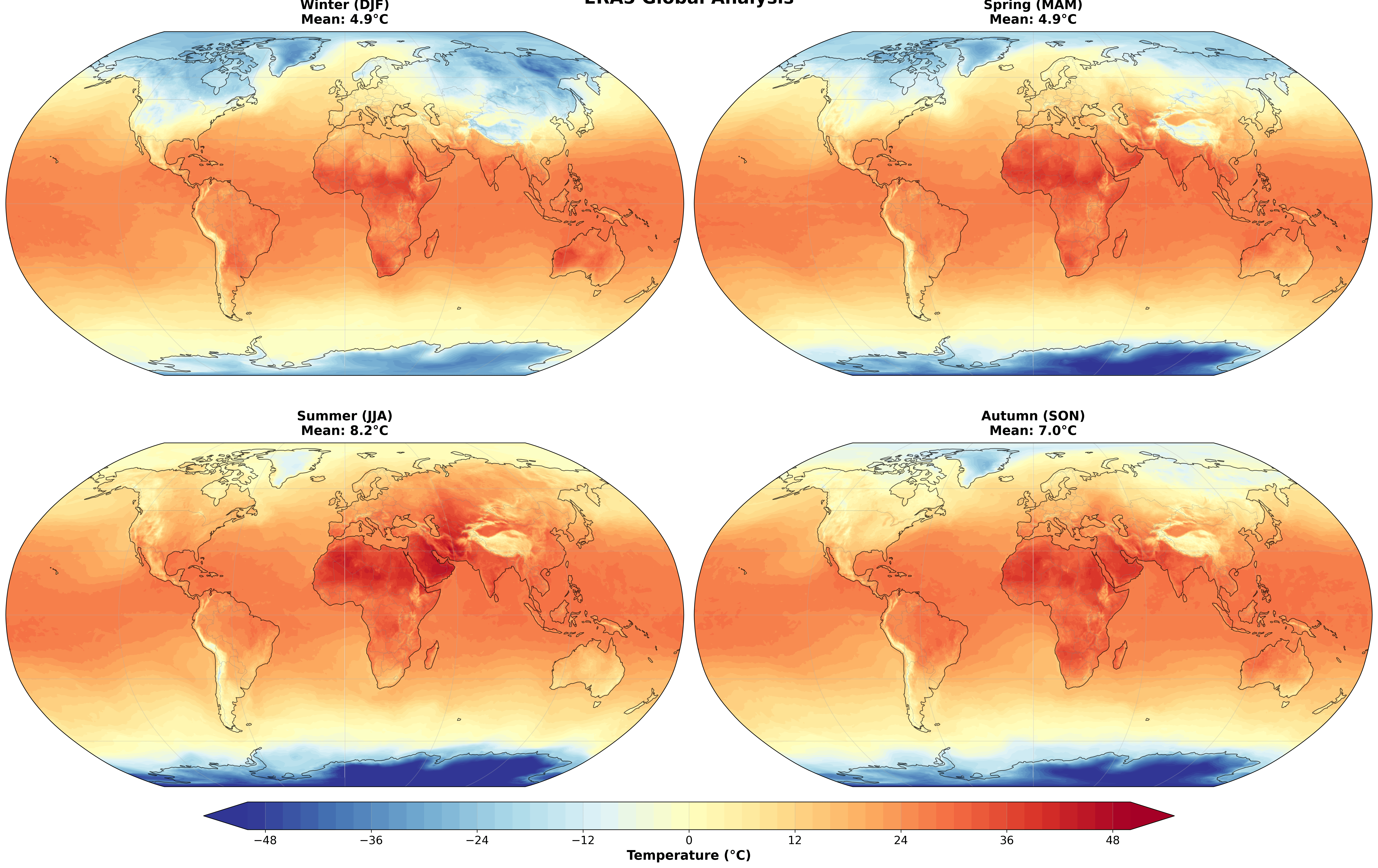}
\end{tabular}
\caption{CMIP6/ERA5 temperature analysis for NYC: historical reanalysis and multi-model SSP2-4.5 projections with ensemble uncertainty.}
\label{fig:sim-study}
\end{figure*}

To demonstrate simulation data handling and generalization beyond observational datasets, we tasked AutoClimDS with analyzing future temperature projections for New York City using CMIP6 and ERA5 datasets. Given only a natural language prompt, the system autonomously: (1) queried the KG for relevant CMIP6/ERA5 experiments, (2) retrieved multi-model ensemble data via authenticated ESGF and CDS API access, (3) performed spatial subsetting to NYC coordinates via geocoding, (4) computed ensemble means and uncertainty ranges, (5) generated comparison visualizations (Fig.~\ref{fig:sim-study}). This workflow required distinct capabilities from observational sea level analysis, specifically multi-model ensemble handling, demonstrating the KG's generalizability across observational and simulation data modalities. GPT-5.1 without KG guidance failed on this task.

\section{OPEN SCIENCE}

Code, data, and documentation are available in supplementary materials \cite{autoclimds}. Resources include KG schema/seed entries, agent workflows, data access scripts, documentation, and tutorials. Users can reproduce experiments and adapt workflows. Importantly, all agent-generated Python codes (via CodeExecutionTool) were logged and preserved, allowing users to inspect preprocessing operations, learn data science workflows, and understand how analyses were constructed, supporting educational use cases beyond pure automation. The modular design enables extensibility through new KG entries (datasets, workflows, ontologies), tools, and documentation. We envision this work as a foundation for collaborative growth by climate scientists, data scientists, educators, students, and citizen science communities.

\section{CONCLUSION}

We built an MVP (Minimum Viable Product) demonstrating that a well-curated knowledge graph enables highly capable AI agents for climate data science workflows, substantially lowering barriers for non-technical users. The claim that "\textit{a knowledge graph is all you need}" reflects a fundamental insight: while agentic systems require LLMs, tools, and infrastructure, the KG provides the irreplaceable structural foundation that makes autonomous scientific reasoning possible. Without procedural knowledge of KG: knowledge, executable access paths, authentication protocols, variable mappings, preprocessing workflows, frontier LLMs hallucinate datasets, select inappropriate sources, and fail to construct workflows, as the GPT-5.1 baseline demonstrates. The KG transforms general-purpose AI into domain-competent scientific agents by encoding the structured memory and reasoning constraints that pure retrieval or LLM reasoning alone cannot provide. In this sense, the KG is the essential enabling component: integrate it with existing technologies (cloud APIs, LLMs, sandboxed execution), and autonomous climate data science becomes achievable. The KG serves as an extensible memory and a unifying reasoning layer across tools/datasets, aligning with cloud data science solutions while creating space for community contributions.

Beyond technical capabilities, this work highlights the broader potential of KGs and AI agents to democratize climate data science. Lowering technical barriers opens opportunities for policy, education, and citizen science participation, while modular cloud-native design ensures scalability and interoperability. The KG provides a foundation for evolving, community-driven commons growing with new datasets, tools, and domain knowledge, positioning it as technical and socio-technical infrastructure for collaborative science. This approach provides pathways toward integrating advanced reasoning capabilities, fostering reproducibility, and accelerating discovery through human–AI partnerships in climate and beyond.

\bibliographystyle{IEEEtran}
\bibliography{climate-references}

@article{righi2020earth,
  title={Earth System model evaluation tool (ESMValTool) v2. 0--technical overview},
  author={Righi, Mattia and Andela, Bouwe and Eyring, Veronika and Lauer, Axel and Predoi, Valeriu and Schlund, Manuel and Vegas-Regidor, Javier and Bock, Lisa and Br{\"o}tz, Bj{\"o}rn and de Mora, Lee and others},
  journal={Geoscientific Model Development},
  volume={13},
  number={3},
  pages={1179--1199},
  year={2020},
  publisher={Copernicus GmbH}
}

@article{eyring2024pushing,
  title={Pushing the frontiers in climate modelling and analysis with machine learning},
  author={Eyring, Veronika and Collins, William D and Gentine, Pierre and Barnes, Elizabeth A and Barreiro, Marcelo and Beucler, Tom and Bocquet, Marc and Bretherton, Christopher S and Christensen, Hannah M and Dagon, Katherine and others},
  journal={Nature Climate Change},
  pages={1--13},
  year={2024},
  publisher={Nature Publishing Group UK London}
}

@article{gettelman2022future,
  title={The future of Earth system prediction: Advances in model-data fusion},
  author={Gettelman, Andrew and Geer, Alan J and Forbes, Richard M and Carmichael, Greg R and Feingold, Graham and Posselt, Derek J and Stephens, Graeme L and Van den Heever, Susan C and Varble, Adam C and Zuidema, Paquita},
  journal={Science Advances},
  volume={8},
  number={14},
  pages={eabn3488},
  year={2022},
  publisher={American Association for the Advancement of Science}
}

@misc{nasaearthdata,
	title = {Open {Science} {\textbar} {NASA} {Earthdata}},
	url = {https://www.earthdata.nasa.gov/about/open-science},
}

@article{ceccato2012data,
  title={Data discovery, access and retrieval},
  author={Ceccato, P and Maxwell, S and Rommel, RG and Jacquez, GM and Benedict, KK and Morain, SA and Yang, P and Huang, Q and Golden, ML and Chen, RS and others},
  journal={Environmental Tracking for Public Health Surveillance},
  pages={229},
  year={2012},
  publisher={CRC Press}
}

@inproceedings{shum2017harvesting,
  title={Harvesting NASA's Common Metadata Repository (CMR)},
  author={Shum, Dana and Durbin, Chris and Norton, James and Mitchell, Andrew},
  booktitle={American Geophysical Union (AGU) 2017 Fall Meeting},
  number={IN51A-0003},
  year={2017}
}

@online{workersAI,
  author  = {Johan Moreno},
  title   = {In a Real-World Study, AI Boosts Worker Productivity by 14\%},
  year    = {2023},
  url     = {https://www.forbes.com/sites/johanmoreno/2023/04/25/in-a-real-world-study-ai-boosts-worker-productivity-by-14/},
  note    = {Forbes, April 25, 2023}
}

@article{meehl2000coupled,
  title={The coupled model intercomparison project (CMIP)},
  author={Meehl, Gerald A and Boer, George J and Covey, Curt and Latif, Mojib and Stouffer, Ronald J},
  journal={Bulletin of the American Meteorological Society},
  volume={81},
  number={2},
  pages={313--318},
  year={2000},
  publisher={JSTOR}
}

@article{williams2011earth,
  title={The Earth System Grid Federation: Software framework supporting CMIP5 data analysis and dissemination},
  author={Williams, Dean N and Taylor, Karl E and Cinquini, Luca and Evans, Ben and Kawamiya, Michio and Lautenschlager, Michael and Lawrence, Bryan and Middleton, Don and ESGF, Contributors},
  journal={ClIVAR Exchanges},
  volume={56},
  number={2},
  pages={40--42},
  year={2011},
  publisher={CLIVAR}
}

@inproceedings{odaka2019pangeo,
  title={The Pangeo ecosystem: interactive computing tools for the geosciences: benchmarking on HPC},
  author={Odaka, Tina Erica and Banihirwe, Anderson and Eynard-Bontemps, Guillaume and Ponte, Aurelien and Maze, Guillaume and Paul, Kevin and Baker, Jared and Abernathey, Ryan},
  booktitle={Annual Workshop on HPC User Support Tools},
  pages={190--204},
  year={2019},
  organization={Springer}
}

@article{raskin2005knowledge,
  title={Knowledge representation in the semantic web for Earth and environmental terminology (SWEET)},
  author={Raskin, Robert G and Pan, Michael J},
  journal={Computers \& geosciences},
  volume={31},
  number={9},
  pages={1119--1125},
  year={2005},
  publisher={Elsevier}
}

@article{wu2022linkclimate,
  title={LinkClimate: An interoperable knowledge graph platform for climate data},
  author={Wu, Jiantao and Orlandi, Fabrizio and O’Sullivan, Declan and Dev, Soumyabrata},
  journal={Computers \& Geosciences},
  volume={169},
  pages={105215},
  year={2022},
  publisher={Elsevier}
}

@article{zhou2020geolink,
  title={Geolink data set: A complex alignment benchmark from real-world ontology},
  author={Zhou, Lu and Cheatham, Michelle and Krisnadhi, Adila and Hitzler, Pascal},
  journal={Data Intelligence},
  volume={2},
  number={3},
  pages={353--378},
  year={2020},
  publisher={MIT Press One Rogers Street, Cambridge, MA 02142-1209, USA journals-info~…}
}

@article{webersinke2021climatebert,
  title={Climatebert: A pretrained language model for climate-related text},
  author={Webersinke, Nicolas and Kraus, Mathias and Bingler, Julia Anna and Leippold, Markus},
  journal={arXiv preprint arXiv:2110.12010},
  year={2021}
}

@inproceedings{green2018opencypher,
  title={openCypher: New Directions in Property Graph Querying.},
  author={Green, Alastair and Junghanns, Martin and Kie{\ss}ling, Max and Lindaaker, Tobias and Plantikow, Stefan and Selmer, Petra},
  booktitle={EDBT},
  pages={520--523},
  year={2018}
}

@inproceedings{yao2023react,
  title={React: Synergizing reasoning and acting in language models},
  author={Yao, Shunyu and Zhao, Jeffrey and Yu, Dian and Du, Nan and Shafran, Izhak and Narasimhan, Karthik and Cao, Yuan},
  booktitle={International Conference on Learning Representations (ICLR)},
  year={2023}
}

@misc{LangGraph,
  title = {LangGraph, built by LangChain Inc},
  author ="LangChain",
  howpublished = {\url{https://langchain-ai.github.io/langgraph/}},
}

@misc{aws_bedrock_claude_sonnet_4,
  title = {Anthropic's Claude in Amazon Bedrock},
  author = {{Amazon Web Services} and {Anthropic}},
  howpublished = {\url{https://aws.amazon.com/bedrock/anthropic/}},
  year = {2025},
  publisher = {{Amazon Web Services}},
}

@online{AWSNeptuneAnalytics,
  author = {{Amazon Web Services}},
  title = {{Amazon Neptune Analytics User Guide}},
  year = {2025},
  url = {https://docs.aws.amazon.com/neptune-analytics/latest/userguide/what-is-neptune-analytics.html}
}

@article{braneonNPCC4NewYork2024,
	title = {{NPCC4}: {New} {York} {City} climate risk information 2022—observations and projections},
	volume = {1539},
	copyright = {© 2024 The Author(s). Annals of the New York Academy of Sciences published by Wiley Periodicals LLC on behalf of The New York Academy of Sciences.},
	issn = {1749-6632},
	shorttitle = {{NPCC4}},
	doi = {10.1111/nyas.15116},
	language = {en},
	number = {1},
	journal = {Annals of the New York Academy of Sciences},
	author = {Braneon, Christian and Ortiz, Luis and Bader, Daniel and Devineni, Naresh and Orton, Philip and Rosenzweig, Bernice and McPhearson, Timon and Smalls-Mantey, Lauren and Gornitz, Vivien and Mayo, Talea and Kadam, Sanketa and Sheerazi, Hadia and Glenn, Equisha and Yoon, Liv and Derras-Chouk, Amel and Towers, Joel and Leichenko, Robin and Balk, Deborah and Marcotullio, Peter and Horton, Radley},
	year = {2024},
	pages = {13--48},
}

@article{acharya_agentic_2025,
	title = {Agentic {AI}: {Autonomous} {Intelligence} for {Complex} {Goals}—{A} {Comprehensive} {Survey}},
	volume = {13},
	issn = {2169-3536},
	shorttitle = {Agentic {AI}},
	journal = {IEEE Access},
	author = {Acharya, Deepak Bhaskar and Kuppan, Karthigeyan and Divya, B.},
	year = {2025},
	pages = {18912--18936},
}

@misc{noaa_onestop,
  author = {{NOAA}},
  title = {{OneStop: A geospatial search platform for environmental data discovery}},
  year = {2023},
  howpublished = {\url{https://www.ncei.noaa.gov/products/onestop}},
  note = {National Centers for Environmental Information}
}

@article{hersbach2020era5,
  title={The ERA5 global reanalysis},
  author={Hersbach, Hans and Bell, Bill and Berrisford, Paul and Hirahara, Shoji and Hor{\'a}nyi, Andr{\'a}s and Mu{\~n}oz-Sabater, Joaqu{\'\i}n and Nicolas, Julien and Peubey, Carole and Radu, Raluca and Schepers, Dinand and others},
  journal={Quarterly Journal of the Royal Meteorological Society},
  volume={146},
  number={730},
  pages={1999--2049},
  year={2020},
  publisher={Wiley Online Library},
  doi={10.1002/qj.3803}
}

@article{eyring2016cmip6,
  title={Overview of the Coupled Model Intercomparison Project Phase 6 (CMIP6) experimental design and organization},
  author={Eyring, Veronika and Bony, Sandrine and Meehl, Gerald A and Senior, Catherine A and Stevens, Bjorn and Stouffer, Ronald J and Taylor, Karl E},
  journal={Geoscientific Model Development},
  volume={9},
  number={5},
  pages={1937--1958},
  year={2016},
  publisher={Copernicus GmbH},
  doi={10.5194/gmd-9-1937-2016}
}

@misc{autoclimds,
  title        = {AutoClimDS: Climate Data Science Agentic AI — Code and Supplementary Materials},
  author       = {Ahmed Jaber and Ayon Roy and Wangshu Zhu and Candace Agonafir and Linnia Hawkins and Tian Zheng and Karthick Jayavelu and Justin Downes and Sameer Mohamed},
  year         = {2026},
  howpublished = {\url{https://github.com/Ajaberr/AutoClimDS}},
  note         = {GitHub repository}
}

\end{document}